\documentclass[letterpaper, 10 pt, conference]{ieeeconf}  

\IEEEoverridecommandlockouts                              

\usepackage[english]{babel}

\usepackage{graphicx}
\usepackage{animate}
\usepackage{epsfig} 				
\usepackage{epstopdf}

\usepackage{listings}
\usepackage{color}
\usepackage{nameref}
\usepackage{hyperref}
\usepackage{amsmath}	 			
\usepackage{amssymb}  				

\usepackage{dsfont}			
\usepackage{mathtools}

\usepackage{epigraph}
\usepackage{lscape}
\usepackage[]{nomencl}				
\usepackage{algorithm}
\usepackage{algorithmic}
\usepackage{multicol}
\usepackage{multirow}
\usepackage{etoolbox}

\usepackage{caption}
\usepackage{subcaption}
\usepackage{listings}
\usepackage{wrapfig}

\usepackage{siunitx}

\usepackage{floatflt}

\usepackage{url}

\let\labelindent\relax
\usepackage{enumitem}

\usepackage{diagbox}

\newcommand{\eg}[0]{e.g.}

\newcommand{\email}[1]{\href{mailto:#1}{\nolinkurl{#1}}}

\newcommand{\link}[1]{\colora{\url{#1}}}

\newcommand{\fig}[1]{Fig.~\ref{#1}}

\newcommand{\tab}[1]{Table~\ref{#1}}

\newcommand{\website}[0]{\url{https://github.com/facebookresearch/pytouch}}
\newcommand{\framework}[0]{PyTouch}
\newcommand{\digit}[0]{DIGIT}
\newcommand{\titlepaper}[0]{\framework{}: A Machine Learning Library for Touch Processing}

\newcommand{\citet}[1]{\cite{#1}}
\newcommand{\citep}[1]{\cite{#1}}

\title{\LARGE \bf \titlepaper}

\author{Mike Lambeta$^{1}$, Huazhe Xu$^{2}$, Jingwei Xu$^{3}$, Po-Wei Chou$^{1}$, Shaoxiong Wang$^{4}$,\\ Trevor Darrell$^{2}$, and Roberto Calandra$^{1}$
\thanks{$^{1}$ Facebook AI Research, Menlo Park, USA}
\thanks{$^{2}$ University of California, Berkeley, USA}
\thanks{$^{3}$ Shanghai Jiao Tong University, China}
\thanks{$^{4}$ Massachusetts Institute of Technology, USA}
}

\begin{document}

\maketitle
\thispagestyle{empty}
\pagestyle{empty}


\begin{abstract}
	With the increased availability of rich tactile sensors, there is an equally proportional need for open-source and integrated software capable of efficiently and effectively processing raw touch measurements into high-level signals that can be used for control and decision-making.
In this paper, we present \framework{} -- the first machine learning library dedicated to the processing of touch sensing signals.
\framework{}, is designed to be modular, easy-to-use and provides state-of-the-art touch processing capabilities as a service with the goal of unifying the tactile sensing community by providing a library for building scalable, proven, and performance-validated modules over which applications and research can be built upon.
We evaluate \framework{} on real-world data from several tactile sensors on touch processing tasks such as touch detection, slip and object pose estimations.
\framework{} is open-sourced at \website{}.
\end{abstract}


\section{INTRODUCTION}
	
	A fundamental challenge in robotics is to process raw sensor measurements into high-level features that can be used for control and decision-making.
For the sense of vision, the field of computer vision has been dedicated to study provide an algorithmic and programmatic method of understanding images and videos. 
In this field, open-source libraries such as PyTorch~\cite{Paszke2019Pytorch}, CAFFE~\cite{jia2014caffe}, and OpenCV~\cite{opencv_library} have enabled the acceleration of research and collectively brought together commonly used techniques in each perspective domain by providing unified interfaces, algorithms, and platforms. 
PyTorch and CAFFE have enabled researchers to develop and scale neural networks by reducing the amount of ground work required for implementing algorithms such as backpropagation or convolution functions. 
Furthermore, OpenCV provided benefits such as a collection of commonly used, tested and optimized functions. 
These frameworks, libraries and platforms all have an underlying goal in enabling further developments in their respective fields through abstracting lower level processes into building blocks for further applications. 

With the increased availability of rich tactile sensors~\citep{Yuan2017GelSight,Lambeta2020DIGIT,Padmanabha2020OmniTact,Abad2020Visuotactile}, the sense of touch is becoming a new and important sensor modality in robotics and machine learning. 
Sensing the world through touch open exciting new challenges and opportunities to measure, understand and interact with the world around us. 
However, the availability of ready-to-use touch processing software is extremely limited -- this results in a high entry bar for new practitioners that want to make use of tactile sensors, which are forced to implement their own touch processing routines. 
We believe that similarly to computer vision, the availability of open-source and maintained software libraries for processing touch reading would lessens the barrier of entry to tactile based tasks, experimentation, and research in the touch sensing domain. 

To address this challenge we introduce \framework{} -- the first open-source library for touch processing that enables the machine learning and the robotics community to process raw touch data from tactile sensors through abstractions which focus on the experiment instead of the low level details of elementary concepts.
\framework{} is designed to be a high-performance, modular, and easy to use library aiming to provide touch processing functionalities ``as a service'' through a pre-trained model collection capable of kick-starting initial experiments to allowing end applications using new or variations of existing sensors the ability to apply transfer learning in levering the performance of the library. 
The software library modularizes a set of commonly used tactile-processing functions valuable for various down-stream tasks, such as tactile manipulation, slip detection, object recognition based on touch, etc. 
Furthermore, the library aims to standardize the way touch based experiments are designed in reducing the amount of individual software developed for one off experiments by using the \framework{} library as a foundation which can be expanded upon for future experimental and research applications.

While tools such as PyTorch and CAFFE currently exist and can be applied to touch processing, precursor development is required to support the necessary algorithms for the experiment and research needs, \framework{} provides a entry point using such tools specifically catered towards the goals of touch processing. 
\framework{} was designed to support both entry level users beginning their first steps in tactile sensing and power users well versed in this domain. \framework{} aims to generalize to diverse tactile input devices that might differ from each other in many design choices. 
This effort joins the recent efforts to standardize robotics research for better benchmarks and more reproducible results. 
Finally, in hand with the framework, we release a set of pre-trained models which are used by \framework{} in the background for tactile based tasks. 

In this paper, we describe the architectural choices of library and demonstrate some of its capabilities and benefits through several experiments. 
Notably, for the first time we evaluate the performance of a machine learning model trained across different models of vision based tactile sensors, and show that this improves performance compared to models trained on single tactile sensors through the use of \framework{} itself to rapidly prototype these experiments. 

The future facing goal of \framework{} is to create an extendable library for touch processing, analogous to what PyTorch and OpenCV are for computer vision, in a way which allows reliable and rapid prototyping of experiments and applications. 
Additionally, through the use of \framework{} as a set of touch processing building blocks and released pre-trained models we lessen the barrier to entry into touch processing while allowing researchers and experimenters to focus on the end goal of their application through proven algorithms and models supplied by the expansion of this library.
We believe that this would beneficially impact the robotic and machine learning community by enabling new capabilities and accelerate research.
	

\section{RELATED WORK}
\label{sec:related}

	Developed fields in computer science build application and research upon frameworks. There are many frameworks existing today which bring together common paradigms used in their respective domains while providing benefit to the community. 
One example of such library is OpenCV~\cite{opencv_library} for the computer vision community, which is a well known and probably the most widely-used data processing library supporting computer vision topics. 

Frameworks which are applicable to the domain of tactile touch sensing are far and few between or not available through open-source channels. 
Observing public repositories and open-source networks, we see very little open-source work pertaining to building a tactile processing framework. 
Although there is work in isolated instances which provide insight into specific problems such as interfacing with robotic skin and providing interfaces for applications \cite{youssefi2015skinware,youssefi2015real}, there are no works which brings commonly used paradigms within the field of tactile touch sensing together or that provide general purpose tools.
Previous attempts include studying friction models~\cite{culbertson2014penn}, physical property understanding~\cite{gao2016deep, burka2016proton,burka2015toward}, in-hand manipulation, determining physical manipulation characteristics and grasp success. 
\cite{belousov2019building} demonstrates individual capabilities related to tactile touch which allow for robotic learning in providing given forces to an object, determining density and texture through stirring actions, and various types of in hand or arm rotation of objects. 
Other work such as \cite{Calandra2018More} learns robotic grasp success through an end-to-end action-conditional model based off of raw tactile sensor input data.
\cite{garcia2019tactilegcn} introduces TactileGCN which estimates grasp stability through a graph convolutional network. \framework{} aims at being a general purpose and open-source solution that combines useful computational tools and capabilities into a single framework and thereby proposes an agnostic library for tactile sensors regardless of their hardware or sensing modalities. 

Another contribution of \framework{} is to push towards hardware standardization by providing flexible interfaces.
Observing the tactile touch community, we saw common trends amongst the hardware used for acquiring touch input: barrier to entry due to cost, sparseness of available hardware arising from single one-off builds, and lack of reproducibility due to semi-closed or closed sourced designs. 
The DIGIT sensor \cite{Lambeta2020DIGIT} aimed to resolve these problems by providing a low cost, scalable and easily machine manufacturable tactile touch sensor. 
Amongst this, we see a pattern of individualized pieces of work in the tactile sensing field but a lack of unification that brings this work together into a library that future work can build upon. 
\framework{} is a step towards unifying the tactile processing field to create a proven, tested and modular open source software library available to all researchers and practitioners.


\section{A LIBRARY FOR TOUCH PROCESSING} 
\label{sec:approach}

	We now describe the desired specifications considered during the design of our touch processing framework, and the corresponding design choices adopted. 
In addition, we describe the software architecture of \framework{}, and several of its key features.

\begin{figure*}[t] 
    \centering
    \includegraphics[width=0.90\textwidth]{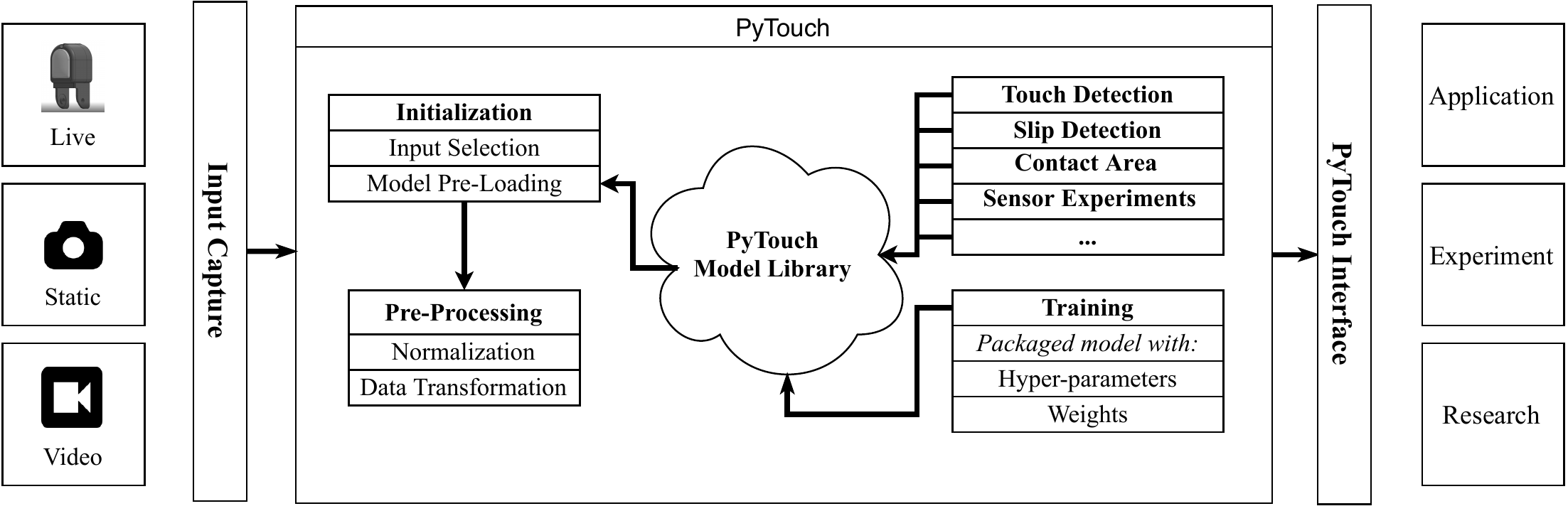}
    \caption{\framework{} high level architecture where tactile touch processing is delivered to the end application "as a service" through released pre-trained models.}
    \label{fig:pytouch-architecture}
\end{figure*}

\subsection{Desired Specifications}
\framework{} aims to provide the same function in the domain of tactile touch processing through the introduction of an open-source library in which proven, tested and performance validated modules allow for scaling touch based applications, research and experimentation. 
This work aims to set forth a need for the unification of tactile processing based software which not only strengthens the field but also decreases the barrier to entry in designing applications and research which utilize tactile touch. 
As such, the main goals of \framework{} are to provide a library that is powered by pre-trained models through the "as a service" paradigm, enables transfer learning through the \framework{} for extending models for new hardware sensors, and deploys joint models which work across a variety of vision based tactile sensors. 

By centralizing touch processing tasks into a software library and through the models provided through \framework{}, this provides a path forward for sensor generalization. 
This is important due to the variance in produced hardware, image transfer layers and mechanical construction of lab-built sensors. 

\framework{} is designed in mind to cater to first time and power users of the tactile touch community. 
High level abstractions of common goals are provided such as, but not limited to: “am I touching an object”, “is the object slipping”, or “what is the touch contact area of the object”. 
Diving deeper, the library can be used to perform analysis techniques such as augmenting the input sensor data, modifying the models used for inference, providing motion and optical flow outputs, or transfer learning techniques for extending the library to non-vision based sensors. 

\subsection{\framework{} Architecture}
\textbf{Pre-trained Models.} \framework{} is built upon a library of pre-trained models which provide real-time touch processing functionalities. The pre-trained models are downloaded after initializing the library with the desired sensor, an example follows in Lst.~\ref{lst:init}. 
Currently, the library supports the DIGIT~\citep{Lambeta2020DIGIT}, OmniTact~\citep{Padmanabha2020OmniTact} and GelSight~\citep{Yuan2017GelSight} sensors, but the 

\begin{lstlisting}[language=Python, caption=Instantiating \framework{} with DIGIT sensor to detect touch and slip. \framework{} reduces the amount of code required to do advanced tasks such as these., label=lst:init]
import pytouch as pt

digit_pt = pt.init(pt.sensors.DigitSensor, tasks={pt.tasks.TouchDetect, pt.tasks.SlipDetect})

digit = ... # Connect and initialize DIGIT sensor
touch, certainty = digit_pt.is_touched(digit.get_frame())
slipping, _ = digit_pt.is_slipping(digit.get_frames())
print(f"Touching object: {touch}, with certainty: {certainty}; Object Slipping: {slipping},")
\end{lstlisting}

library can be used to train models using data from other vision or non-vision based tactile sensors. 
At the current state, \framework{} provides the following functions: contact classification, slip detection, contact area estimation, and interfaces for training and transfer learning. By developing this library with modularity and scalability in mind, we plan to extend features and provide new models "as a service" to the community and supporting new additions from the community itself. 
Additionally, deploying these models "as a service" aids in the experimentation of different approaches to the same task by swapping pre-trained models for another without changing high-level software. 
This allows for performance benchmarking of real-world experiments for the purposes of creating a tactile task baseline.

\textbf{Sensor Abstraction.} \framework{} can be used in a number of ways such as, providing pre-captured video streams, static images, or live video captured from an input device. The library is instantiated with a known hardware device which in the background provides inference and functionality based on the hardware specific pre-trained models. The set of supported devices is extendable through the library as well, and they may be initialized by supplying a custom configuration file with models respective to that device. \framework{} provides the training routines for each library capability. Once trained, the weights can then be used through \framework{} for different hardware input devices without changing the application software. 

\textbf{Tactile Task Abstractions.} \framework{} provides implementations of commonly used features in the tactile touch domain. The aim of the library is to reduce friction when experimenting with hardware implementations and robotic workspaces utilizing touch. Two interfaces are provided in initializing the library for use in applications. The first interface, Lst. \ref{lst:init}, abstracts all internal modules into a single module while the second interface allows for individual module level access to each task for power users.

With each interface, an input device is registered which loads the appropriate pre-trained models. These pre-trained models are downloaded during initialization or when the appropriate class is instantiated in the background. The registered input device provides defaults as to which pre-trained models to use, and defines the internal data transformation to the models. Additionally, \framework{} supports extending the input devices for custom hardware through adding an additional input device module, or by specifying the custom device module which allows the user to provide custom pre-trained models.

\framework{} is built upon a proven and widely used machine learning framework, PyTorch~\cite{Paszke2019Pytorch}. 

\textbf{Model and Network Agnostic.} One of the crucial benefits of \framework{} is the abstraction of high-level function calls which allows end-users to integrate their own models specific to experiments, sensors, or use our pre-trained models and perform fine-tuning. \framework{} introduces standardization of learning-based methods through the application programming interface of being able to swap models as needed, or change them through initialization parameters. \framework{} abstracts the low level details in order to allow new users to experiment through models served by the \framework{} team, or through the community. This ultimately allows \framework{} to be agnostic to model support in supporting conventional deep CNN models and through non-conventional learning based methods. Furthermore, this abstraction enables new users to touch processing to begin experimentation without having knowledge of model, network and parameter design choices but rather to focus on the experimentation itself.


\section{EXPERIMENTAL RESULTS}
\label{sec:result}

	We now provide several examples which outline the advantages in using the \framework{} library. 
Predominantly, the flexibility of the library to work with different vision-based tactile sensors such as DIGIT~\citep{Lambeta2020DIGIT}, OmniTact~\citep{Padmanabha2020OmniTact} and GelSight~\citep{Yuan2017GelSight}. 
Secondly, the inherent design of the library to enable capabilities through the "as a service" paradigm. 
\framework{} releases pre-trained models which are tested, validated and supports additions via the community to extend library capabilities. 

\subsection{Touch Detection}
A first functionality that \framework{} provides is touch detection. 
This is a basic functionality which is often used as a subroutine in more complicated manipulation sequences.
However, training high-accuracy touch detection models can be difficult and require extensive data collections.
Here we show that using one \digit{} sensor, we can determine if an object is being touched across multiple variations of that same sensor. 
The robustness of the library depends on the generalization of the model and due to manufacturing variances in the sensors and image transfer layers, no one sensor is exactly the same.

\textbf{Formulation.} Touch detection is formulated as an image classification task where given an input image $\mathbf{X}$, we assign a label $y$ which is $0$ for data without touch and $1$ for data with registered touch.

\begin{figure}[t]
    \centering
    \includegraphics[width=\linewidth]{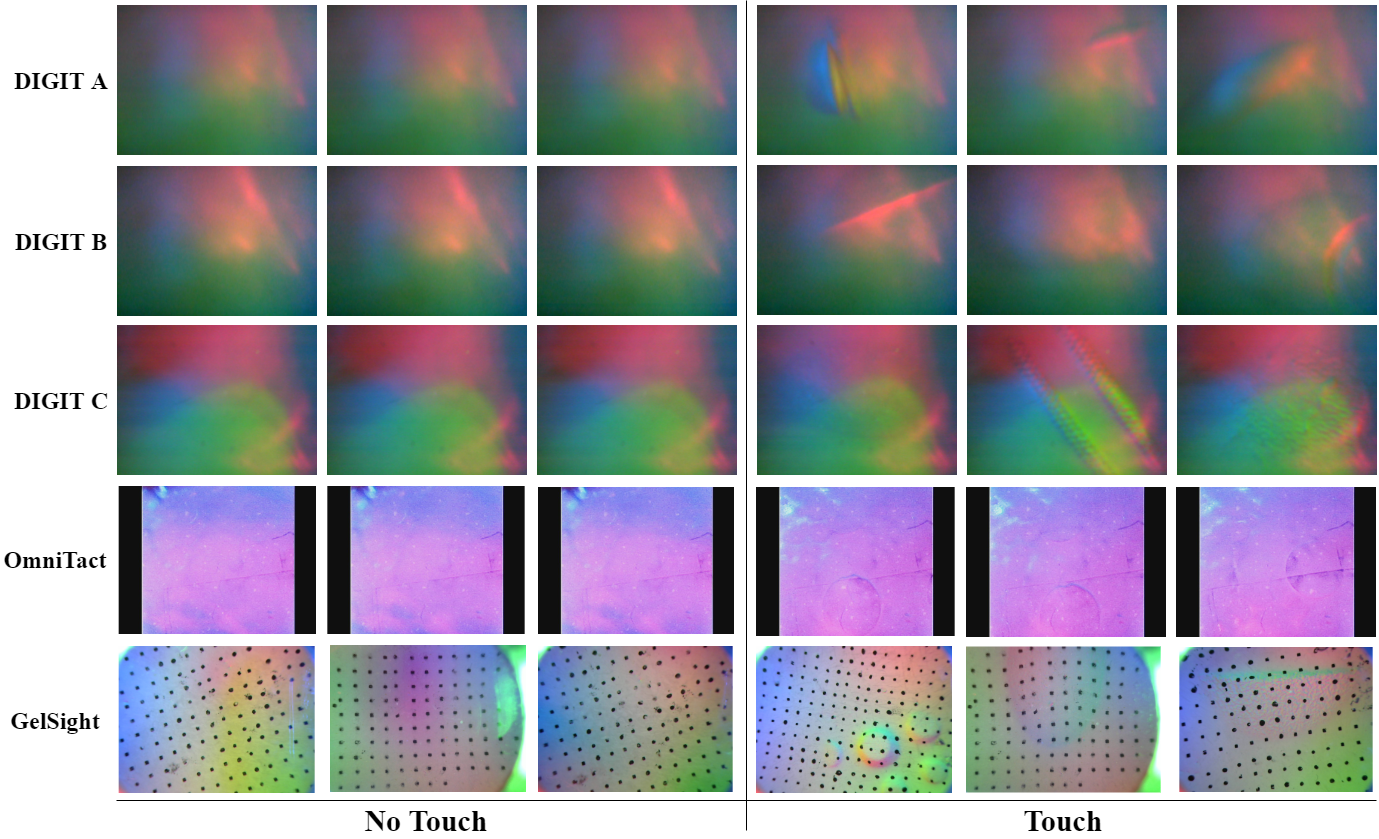}
    \caption{Examples of data used to train our touch prediction models. The dataset include data across several DIGITs, OmniTact, and GelSight sensors showing different lightning conditions and objects of various spatial resolutions.}
    \label{fig:exp-digit-touch-detect}
\end{figure}

\textbf{Datasets.} 
The dataset for DIGITs was collected touching random household and office objects of various spatial resolutions.
This dataset was split by unique device serial numbers in contact, with over overall 2278 samples from 3 different sensors.
The OmniTact dataset was provided by the authors of \citep{Padmanabha2020OmniTact} and consisted of 1632 random presses of a single object. 
The GelSight samples were extracted from the dataset provided in \cite{Calandra2018More}, which consisted of before and after grasp images from two GelSight sensors. 
We used only a subset of the total dataset for a total of approximately 1000 images. 
The input images collected for DIGIT are images with size $240 \times 320 \times 3$, OmniTact images are $480 \times 640 \times 3$ and GelSight images are $1280 \times 960 \times 3$.

\textbf{Training and Architecture.} The touch detection capability is based off of a pre-trained ResNet-18 model. 
Two modalities are used for training, the first is where images are provided to the model without a reference and the second is where the image is concatenated with a reference no-touch image. The reference is unique to each sensor which is a frame where the sensor is not touching any object. 
The input data is normalized and down-sampled to $64 \times 64$. 
The data was split to ensure an equal distribution of device serial numbers. For training with a reference no-touch image, a unique reference image was provided to the model and concatenated to each input image. 
Finally, a joint dataset was created to include all devices in order to show model robustness and generalization.

\begin{table}[t]
\centering
\resizebox{\columnwidth}{!}{%
\begin{tabular}{|l|ccc|}
\hline
\backslashbox{Model~~~~~~~~}{Sensor} & DIGIT~\cite{Lambeta2020DIGIT}& OmniTact~\cite{Padmanabha2020OmniTact} & GelSight~\citep{Yuan2017GelSight}\\ \hline
Single models (no reference) & $95.5 \pm 1.2$                                    & \multicolumn{1}{|c|}{$98.4 \pm 1.1$}   & $93.7 \pm 1.2$\\ 
Single models (with reference)    & $95.8 \pm 1.3$                                    & \multicolumn{1}{|c|}{$98.5 \pm 1.1$}    & N/A\\
\hline
Joint model (no reference)  & $96.1 \pm 1.0$                                    & $99.1 \pm 0.4$   & $98.3 \pm 0.6$\\ 
Joint model (with reference)     & $96.2 \pm 1.1$                                    & $99.5 \pm 0.3$   & N/A  \\
\hline
\end{tabular}
}

\caption{Classification accuracy [$\%$] of touch detection (mean and  std) using cross-validation ($k=5$). The joint models are trained with data from all three sensors: DIGIT, OmniTact and GelSight.}
\label{tab:touch-detection}
\end{table}

\begin{figure}[t]
    \centering
    \includegraphics[width=0.75\linewidth]{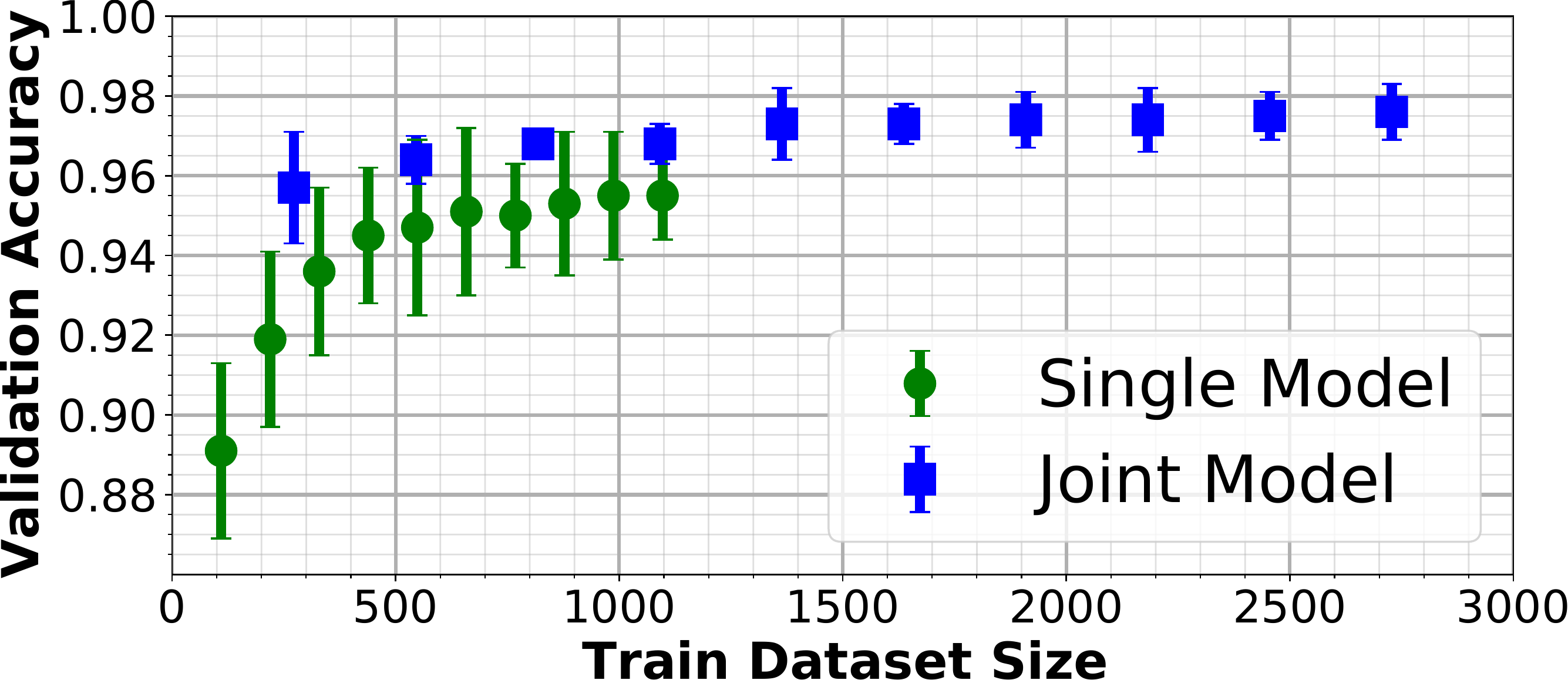}
    \caption{Cross validation ($k=5$) accuracy with varying train dataset size for single and joint models. With the same amount of data, training a joint model using data across multiple sensors (i.e., DIGITs, OmniTact and GelSight) results in better model performance compared to training from a single sensor.}
    \label{fig:exp-constrain-train-dataset-size}
\end{figure}

\begin{figure}[t]
    \centering
    \includegraphics[width=0.80\linewidth]{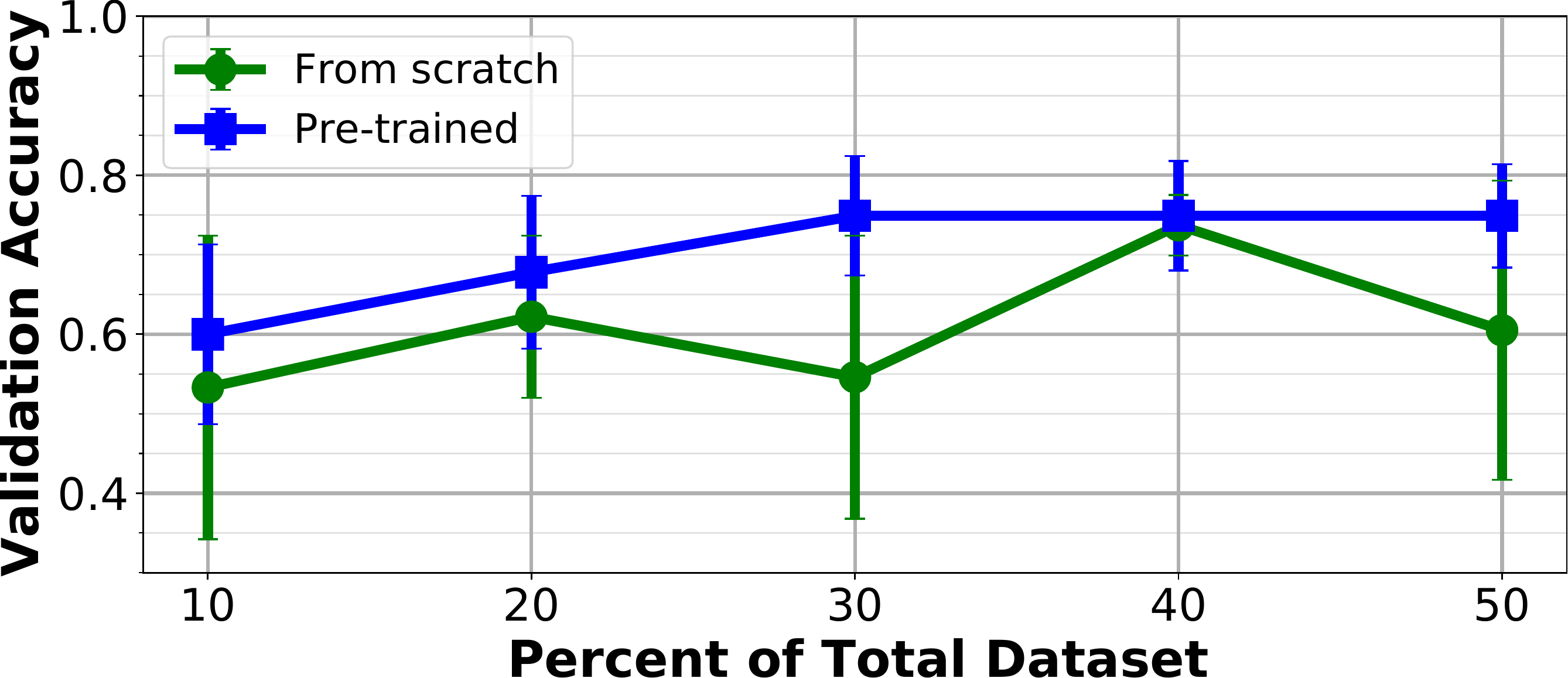}
    \caption{Cross validation ($k=5$) accuracy with varying train dataset size when training a model from scratch or starting from the pre-trained joint model. The validation is performed on dataset of size 482 from a DIGIT sensor. Using the joint model to fine-tune against a new dataset results in better model performance with lower dataset sizes compared to training a the model from scratch for the same wall-clock time of \SI{30}{\second}. Thus \framework{} can provide benefit to end applications when using the "as a service" pre-trained models.}
    \label{fig:exp-fine-tune-exp}
\end{figure}

\textbf{Results and Ablative Study.} A model specific to each sensor and a joint model comprising of all sensor data across each unique sensor was trained. 
We show in \tab{tab:touch-detection} that the joint model performs better across a 5-fold cross validation test. 
The model trained for the OmniTact sensor shows higher performance than the DIGIT model, however, observing the data in the OmniTact dataset, sample images shown in \fig{fig:exp-digit-touch-detect}, we see that the breadth of sample uniqueness is minimal. 
We also show the effects of constraining the train dataset sizes ranging from $10\%$ to $100\%$ of the total dataset size in \fig{fig:exp-constrain-train-dataset-size} that creating a joint model which uses multiple sensors results in better performance while using a similar amount of data in the training dataset. 
To our knowledge, this is the first study on multiple sensors manufacturers and sensor variants to show benefit of a touch processing library. 

\begin{figure*}[t]
    \centering
    \includegraphics[width=0.72\linewidth]{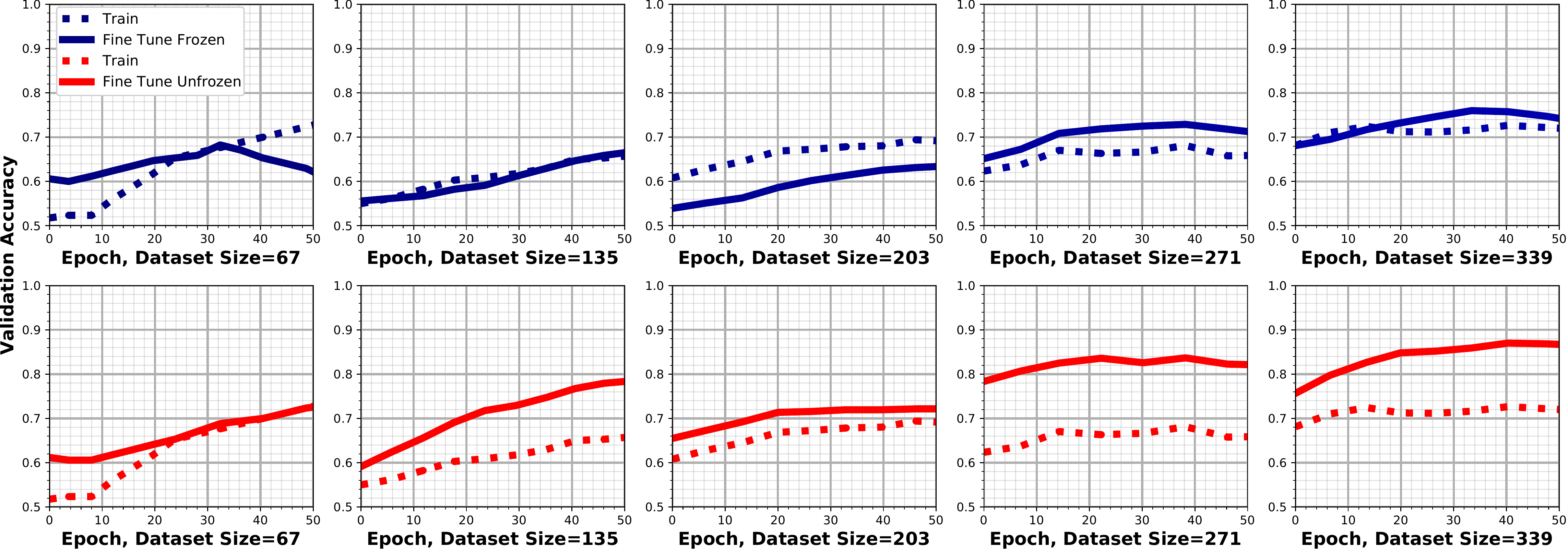}
     \caption{Comparison of model trained just from the dataset collected from the GelSight \citep{Yuan2017GelSight} sensor ("Train") against model trained using the pre-trained joint model and fine-tuning the last 3 layers ("Fine Tune Unfrozen") or just the last layer ("Fine Tune Frozen"). The results show using the pre-trained and fine-tuning the last 3 layers model significantly improve the performance of the models. This suggests that the pre-trained model learned useful features that can generalize to different sensors.}
    \label{fig:exp-fine-not-frozen}
\end{figure*}

Furthermore, we show a direct benefit of using \framework{} in regards to fine-tuning a smaller dataset acquired from a new sensor on the joint model. 
A large dataset was collected from a new sensor and then constrained from $50\%$ to $10\%$ of the initial size.
The results on the touch detection task show that using a pre-trained model can both increase the final performance and the training wall-clock time when using our pre-trained model and then fine-tuning, compared to a model trained exclusively from new data. 
We ablate this experiment over different dataset size to show how the advantages are more pronounced the smaller the dataset used for fine-tuning is. 
In \fig{fig:exp-fine-tune-exp} we show the results of the fine-tuned model compared to the model trained from the dataset alone for the same wall-clock time of \SI{30}{\second}. 
Furthermore, we shown in \fig{fig:exp-fine-not-frozen} that introducing new sensors using \framework{} as a base results in better performance with less samples through transfer learning from the released pre-trained models \framework{} provides. 

The \framework{} library is designed with the goal to support real time experimentation with input into the touch detection model resulting in an average performance of \SI[separate-uncertainty,multi-part-units=single]{5.89 \pm 0.46}{\milli\second}. 
Tactile input sensors are most commonly seen with input capture rates of 24 to 60 frames per second which places each frame between $17$ and \SI{41}{\milli\second}. 
This enables \framework{} for use in real time applications which require high frame rates of up to \SI{140}{fps}.

\subsection{Contact Area Estimation}
\begin{figure}[t]
    \centering
    \includegraphics[width=0.7\linewidth]{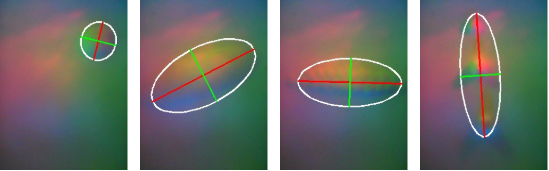}
    \caption{Examples of contact area estimations using 4 objects: glass ball, pen, cable and fork}
    \label{fig:pose_demo}
\end{figure}

When grasping an object with a single or multiple sensor it is useful to know the surface contact area of the object against the sensor. 
We show an example of this in \fig{fig:pose_demo} where a number of object contact areas are estimated with single and multiple point features using a DIGIT~\citep{Lambeta2020DIGIT} tactile sensor. 
The contact area estimated is not referenced to a ground truth, but rather provides the areas of contacts and estimates the direction of contact with respect to the surface of the touch sensor. 

\framework{} provides the contact area estimation capability for many input formats such as raw sensor input, video and static images. Additionally, providing centroid information and estimated size of the semi-major and semi-minor axis of the fitted ellipse. 

\subsection{Library Experimentation}
\begin{figure}[t]
    \centering
    \includegraphics[width=0.75\linewidth]{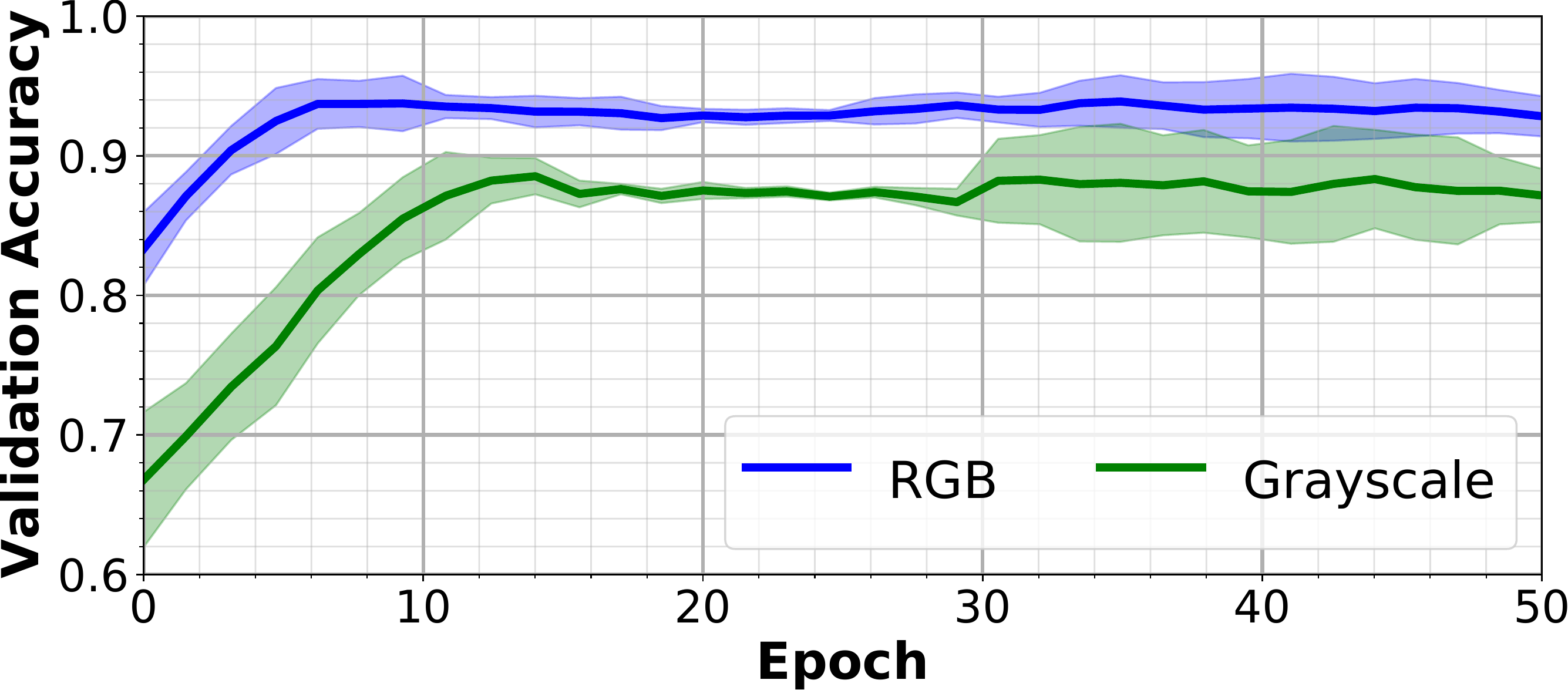}
    \caption{DIGIT \citep{Lambeta2020DIGIT} sensor loss in performance when comparing monochromatic light to RGB lighting}
    \label{fig:color_exp}
\end{figure}

There are a number of parameters available to physical hardware sensor design used in the field of tactile touch sensing and due to the cumbersome nature of hardware development, progress and iteration is often slow and costly. Often, it is valuable to experiment with physical based parameters prior to building a full prototype to reduce the number of iteration cycles. \framework{} proposes a module which allows in the experimentation of these physical based parameters against previous benchmarks obtained by the joint and single sensor models. We show in \fig{fig:color_exp} that for the DIGIT~\citep{Lambeta2020DIGIT} sensor which relies on RGB lighting that a physical hardware change to using monochromatic light results in less performance compared to chromatic light. 

\subsection{Slip Detection}

Slip detection is crucial for many robotics tasks \eg{in-hand manipulation, grasping}. 
In \framework{}, we incorporate a slip detection module as a pre-trained model. 
We also provide APIs for training customized models. 
The module design and implementation are described as follows.

\textbf{Formulation.} We formulate this task as a video classification task: given an image sequence $\mathbf{X}=\{x_1, \cdots, x_T\}$, we assign a label $y$ from a discrete set $\mathbb{Y}$. 
We additionally label the slip starting and ending timestamp $t_s$ and $t_e$ for extracting shorter sub-sequences. 
Specifically, the $x_t$ is an RGB image with size $240 \times 320 \times 3$, the label $y$ is a binary label with $1$ representing slip and $0$ representing non-slip. 
We note that both rotational and translational slip are labeled as slip without distinction.

\textbf{Dataset.} We collected 40 objects with a large spectrum of surface finishes, roughness, and features. 
This can be thought of as ``texture-ness'' from the heavily textured pen holder to an almost textureless plastic bottle. 
For each object, we collected 10 slip sequences and 10 non-slip sequences. Each sequence is consist of 128~frames in total. 
All the data are collected by using DIGIT sensor to slide or touch an object with 30~Hz frequency~\cite{Lambeta2020DIGIT}.

\textbf{Training and Architecture.} We used ResNet-18 and ResNet-18-based 3D convolutional neural network as our architecture. 
For both architecture, we concatenate all the images as input. 
We use cross-entropy loss as the objective of the binary classification task. We follow the standard approach to normalize the input data, downsample the image to $112 \times 112$ and use BatchNorm for better performance. 
We use Adam optimizer with learning rate $0.001$. 
Since it is video classification, we compare the results in two data split setting. The first setting is ``split by sequence`` where we use 20\% of the data from each object as test sequences. 
The second split is ``split by objects'' where we withhold 20\% objects as test data. We note that the latter is much harder than the former due to the need of generalization.

\begin{table}
\centering
\begin{tabular}{llllll}
\hline
 & \multicolumn{2}{l}{Split by Sequences {[}\%{]}} &  & \multicolumn{2}{l}{Split by Objects {[}\%{]}} \\ \cline{2-3} \cline{5-6} 
 & $len = 12$ & $len = 128$ &  & $len = 12$ & $len = 128$ \\ \hline
ResNet-18 & 90.6 & 97.9 &  & 89.2 & 96.3 \\
3D Conv & N/A & 91.7 &  & N/A & 81.2 \\ \hline
\end{tabular}
\caption{Slip detection accuracy averaged over all trials with different architectures, number of frames and dataset split methods.}
\label{tab:slip-sequence-data}
\end{table}

\textbf{Results and Ablative Study.} 
In \tab{tab:slip-sequence-data}, we show the classification accuracy with all 128 frames or 12 frames as inputs. 
We notice that the original ResNet-18~\cite{DBLP:journals/corr/HeZRS15} outperforms 3D conv net. 
We attribute this to the high variance of slip velocity in the dataset that confuse the 3D convolutional kernel. 
We also find that the accuracy on ``split by object'' is lower than ``split by sequence'', which confirms our claim about the requirement for better generalizability in the former split. 
In the same table, we also extract a shorter sequence with 12 frames (approximately 400 milliseconds) as in \cite{li2018slip} from the starting timestamp $t_s$ to $t_{s+12}$. We find the performance drops slightly to around 90\%.


\section{CONCLUSION}
\label{sec:conclusion}

	As the field of touch sensing evolves, a need arises for advanced touch processing functionalities capable of efficiently extracting information from raw sensor measurements.
Prior work provided some of these functionalities, but typically ad-hoc -- for single sensors, single applications, and not necessarily with reusable code (e.g., open-source).
To better support and grow the tactile sensing community, we believe that is necessary the creation of a software library which brings together advancements in the touch processing field through a common interface. 
To address this need, we introduce \framework{}, a unified library for touch processing. 
Our goal with this library is to enable academics, researchers and experimenters to rapidly scale, modify and deploy new machine learning models for touch sensing. 
In this paper, we detailed the design choices and the corresponding software architecture of this library.
\framework{} is designed with open-source at the core, enabling tactile processing "as-a-service" through distributed pre-trained models, and a modular and expandable library architecture. 
We demonstrate several of the feature of \framework{} on touch detection, slip and object pose estimation experiments using 3 different tactile sensors, and we show that a unified tactile touch library provides benefit to rapid experimentation in robotic tactile processing.
We open-source \framework{} at \website, and aim in continuing the evolution of this library to further the field of touch processing.


\addtolength{\textheight}{-12cm}   




\section*{ACKNOWLEDGMENT}

We thank Omry Yadan for insightful discussions and assistance to integrate Hydra~\citep{Yadan2019Hydra}, Frederik Ebert, Stephen Tian and Akhil Padmanabha for providing the OmniTact dataset, and Joe Spisak for helping with community building and project vision.


\clearpage
\bibliographystyle{IEEEtran}
\bibliography{paper}

\end{document}